# Evaluating Vision-Language Models as a Zero-Shot Learning Alternative to You Only Look Once and Optical Character Recognition for Nigerian License Plate Recognition

Ismail Ismail Tijjani
*department of Mechatronics Engineering,*
*Bayero University Kano(BUK).*
Kano, Nigeria.
ismailismailtj@gmail.com

Ahmad Abubakar Mustapaha
*School of Computer Science and Engineering,*
*VIT-AP Uiversity,*
Andhra Pradesh, India.
ahmadmustapha35@gmail.com

Sunusi Ibrahim Muhammad
*department of Petroleum Engineering*
*Bayero University Kano(BUK).*
Kano, Nigeria.
sunusimuhammada@gmail.com

Muhammad Bashir Aliyu
*department of Mechatronics Engineering.*
*Aliko Dangote University of Science and technology, Wudil.*
Kano, Nigeria.
mb25511644@gmail.com

***Abstract*— License Plate Recognition (LPR) systems are critical tools in traffic monitoring, security enforcement, and urban mobility management. Traditional LPR systems often rely on a multi-stage pipeline involving object detection using You Only Look Once (YOLO) and Optical Character Recognition (OCR), which suffer from limitations such as high resource demands, poor performance in unstructured environments, and the need for large annotated datasets. This study explores the potential of Vision-Language Models (VLMs) as a unified, zero-shot learning solution for Nigerian license plate recognition. Using a curated dataset of 88 challenging real-world images collected in Nigeria, we evaluate five selected VLMs: Gemini 2.0 Flash Exp (Google DeepMind), Qwen2.5-VL-7B-Instruct (Alibaba), GPT-4o (OpenAI), Claude 4 Sonnet (Anthropic), and Llama 3.2 Vision 90b (Meta). Results based on Character Error Rate (CER) reveal that Gemini and Qwen significantly outperform other models in both accuracy and robustness, on the challenging image scenarios. This work highlights the practical advantages of VLMs over YOLO+OCR, questions the claims by model providers, and compares the performances of the VLMs.**

## I. INTRODUCTION

Security remains a major global concern that nations across the world are striving to address. With the rise in threats to public safety, governments and institutions are continuously seeking reliable and efficient solutions to enhance their security infrastructure. The rapid advancement of Artificial Intelligence (AI) in recent years has opened new opportunities for tackling security challenges through automation and intelligent decision-making systems.

Transportation and traffic regulation are critical components of a secure society. Roads are the most widely used infrastructure connecting nations, states, communities, and regions. Ensuring order, safety, and monitoring in transportation systems plays a vital role in national security efforts. Among various technologies developed for this purpose, LPR systems have proven to be effective in vehicle tracking, law enforcement, access control, and surveillance [16]. LPR systems have evolved over the years, typically consisting of two main tasks: detection (locating the license plate in an image) and extraction (recognizing the alphanumeric characters). The current state-of-the-art (SOTA) solutions often use YOLO models for detection and OCR systems for text extraction.

However, the traditional YOLO+OCR pipeline presents several limitations, including high resource demands for dataset collection and annotation, model training, reliance on expensive multi-stage infrastructure, inconsistent performance under real-world conditions, and the need for continuous maintenance to adapt to evolving data as shown in Fig. 1.

To address these limitations, this study investigates the performance of five VLMs in a zero-shot learning setting using a single, well-structured prompt that enables both detection and extraction in one pass as shown in Fig 2[17]. The evaluation is conducted on a dataset of challenging Nigerian license plates, many of which are low-resolution, partially visible, taken from difficult angles, or contain ambiguous characters, conditions that often challenge even human interpretation. The goal is to determine how well these models generalize to real-world, applications without task-specific fine-tuning. While many VLMs developers claim superior performance based on controlled benchmarks, this work offers a critical, real-world evaluation of those claims by testing the models under diverse and practical conditions [17].

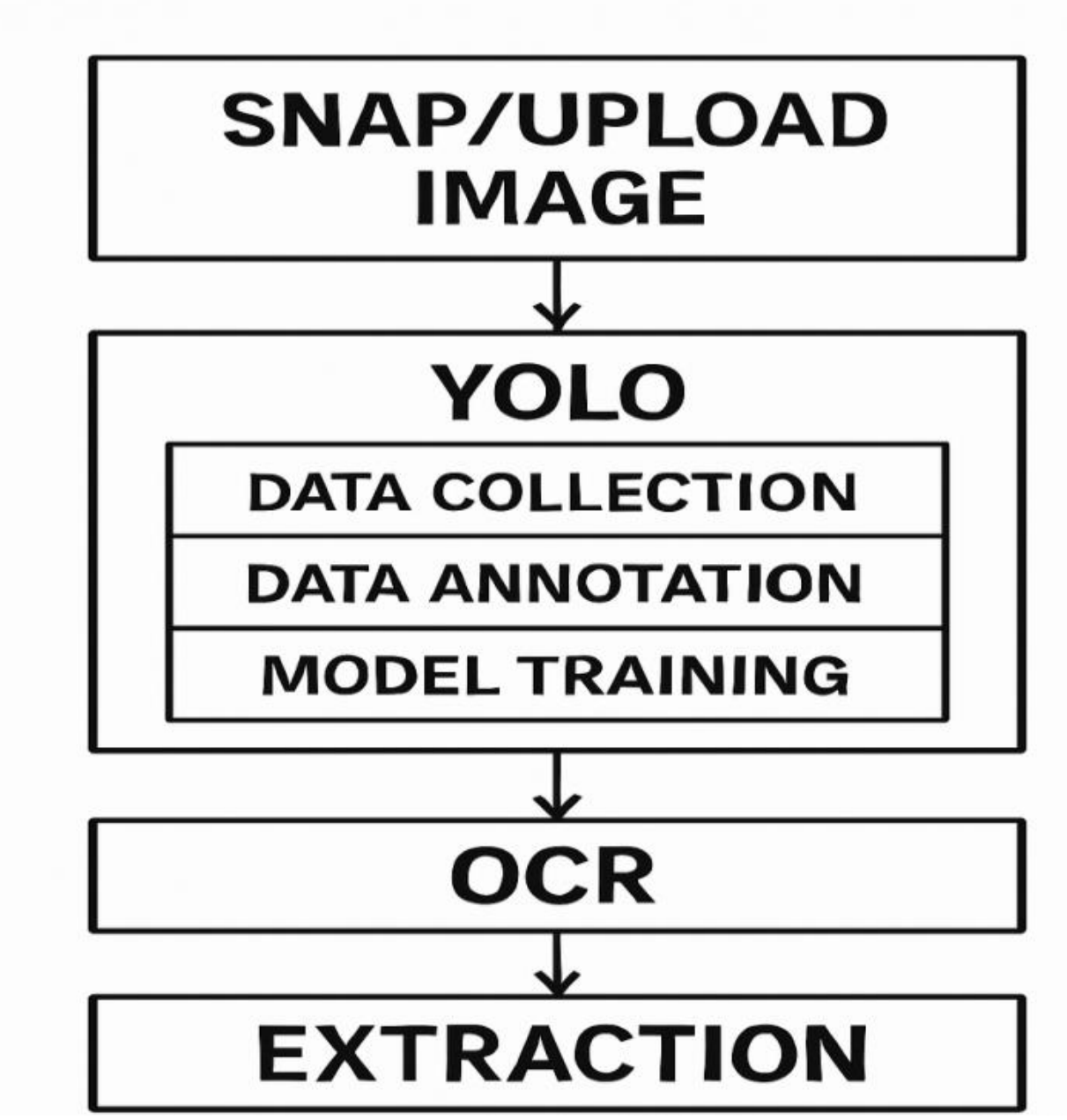

Fig. 1. YOLO+OCR Approach Pipeline

## II. Literature Review

Several approaches have been adopted during the early evolution of AI-based LPR systems, most of which employed a multi-module approach. In 2004, an LPR technique was proposed which is composed of two primary modules: a license plate localization module and a license number identification module. The former, characterized by fuzzy logic techniques, aimed to extract license plates from input images, while the latter employed neural network-based models to identify the characters on the plate [1]. [2] proposed a three-module system consisting of car detection, license plate segmentation, and a Support Vector Machine (SVM)-based character recognition module. Similarly, [3] developed a system comprising plate detection, character segmentation, and recognition modules.

With advancements in deep learning, YOLO-based architectures gained prominence due to their real-time object detection capabilities. Rayson Laroca et al. introduced a robust and efficient LPR system built upon the YOLO object detector, achieving state-of-the-art results [4]. In another study, YOLOv3 was utilized for region of interest (ROI) detection, while a Convolutional Neural Network (CNN) handled OCR [5]. An improved approach was proposed in [6], where vehicle color recognition (VCR) was integrated with license plate detection and OCR using a single-stage YOLO-based multi-task model. This model enhanced efficiency by performing all three tasks concurrently using parallel object detection heads on the YOLO backbone.

The combination of YOLO and OCR has become the de facto standard for LPR tasks. However, several limitations have been identified. For example, [4] reports that performance drops significantly in challenging conditions, such as low-resolution images, excessive noise, poor lighting, or motion blur. Non-standard plates with unusual sizes, fonts, or layouts further hinder accurate recognition. Environmental factors such as glare from sunlight or reflections from bright surfaces can obscure characters, leading to unreliable results [7]. Moreover, [8] highlights the computational burden associated with YOLO+OCR systems, which often require high-end GPUs. This restricts their deployment on edge devices, which are crucial for real-time use cases. Also, these systems often rely on specific camera angles, controlled lighting, and plain backgrounds, making them unreliable in real-world scenarios such as busy, urban roads with complex and unconstrained environments.

Most traditional visual recognition studies rely heavily on crowd-annotated datasets to train Deep Neural Networks (DNNs), with each model typically specialized for a single task. This paradigm is labor-intensive, time-consuming and also limits generalizability across tasks. To address these challenges, VLMs have emerged as powerful alternatives. Trained on vast collections of web-scale image-text pairs, VLMs learn rich cross-modal representations that enable zero-shot transfer across a wide range of visual recognition tasks using a single unified model [9].

VLMs exhibit capabilities in region-level captioning, visual reasoning, classification, and expression comprehension. They can also handle task-guided instruction following and caption generation, making them highly versatile in downstream applications [10]. One of the key enablers of their success is prompt engineering, the process of crafting effective natural language prompts to guide model outputs. As highlighted in [11], the performance of pre-trained VLMs in downstream tasks is strongly influenced by the design of the textual prompt. Through vision-language pre-training, these models align visual and textual information in a shared latent space, which allows them to synthesize task-specific outputs based on natural language descriptions.

Recent advances further highlight the potential of VLMs in OCR tasks. For instance, Ocean-OCR [12] employs the Native Resolution Vision Transformer (NaViT) [13] and has demonstrated superior performance over traditional OCR models like TextIn and PaddleOCR across multiple benchmarks, including DocVQA, ChartQA, TextVQA, and OCRBench. UReader [14] was introduced as a universal OCR-free system for visually-situated language understanding. Instead of relying on conventional pretraining pipelines, UReader fine-tunes a multimodal large language model (MLLM) using diverse visually grounded datasets via cost-efficient instruction tuning [15]. Remarkably, UReader achieved state-of-the-art OCR-free performance on 8 out of 10 tasks across five domains, reinforcing the effectiveness of instruction-tuned MLLMs in text-rich visual environments.

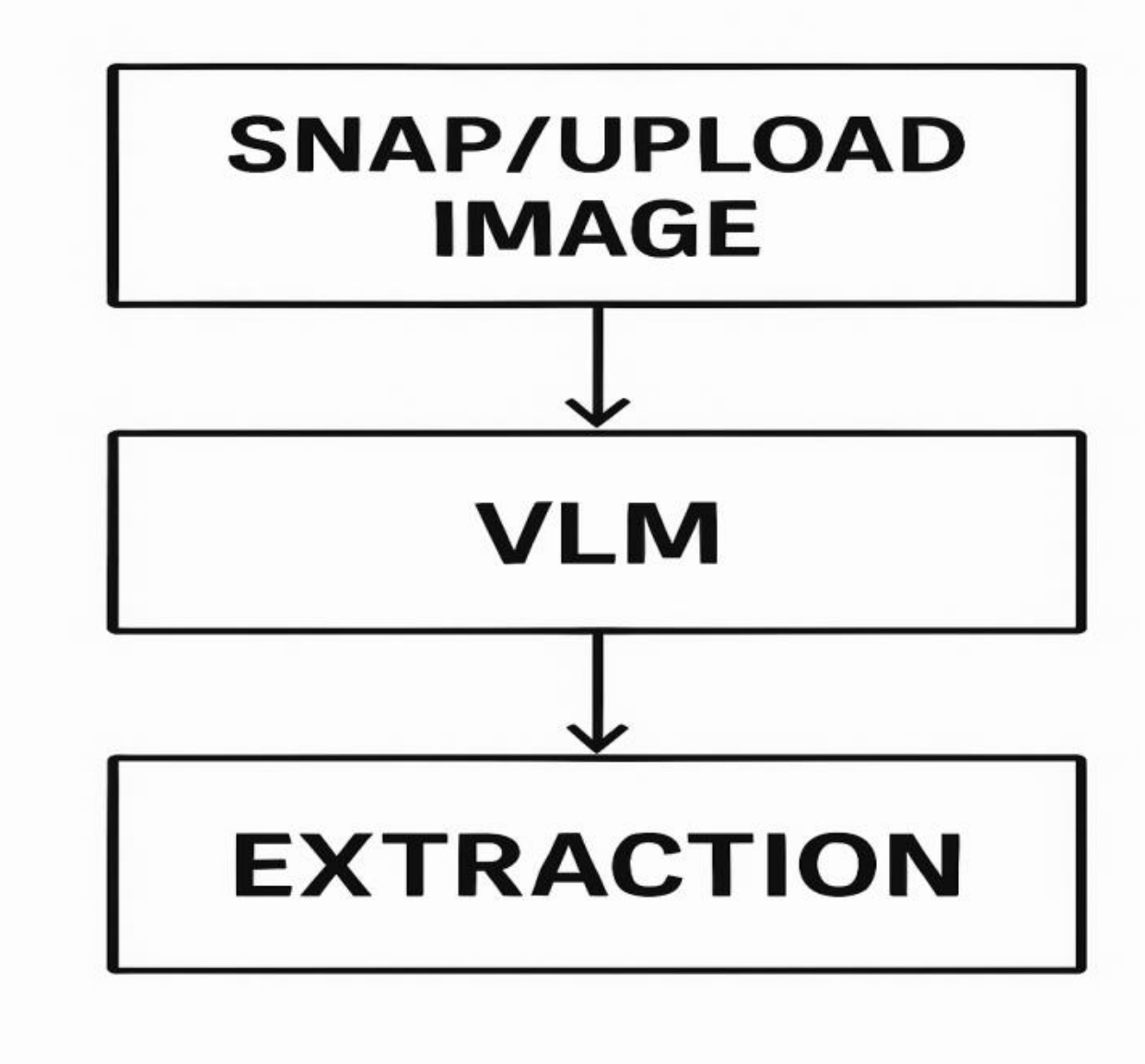

Fig. 2. Proposed VLM Apprach

## III. METHDOLOGY

This study is motivated by the limitations encountered in our prior implementation of Nigerian License Plate Detection and Recognition System, which utilized the YOLO object detection framework in combination with PaddleOCR for alphanumeric text extraction. While that approach demonstrated acceptable performance under controlled conditions, its accuracy significantly declined in more realistic environments.

### *A. Dataset and Collection Process*

The dataset employed in this study is a subset of a larger collection of Nigerian vehicle license plate images curated in the course of our previous research. A total of 88 images samples were selected based on their challenging nature, diversity and potential to evaluate model robustness across unconstrained scenarios. The original dataset was manually acquired by EJAZTECH.AI team across multiple locations in Kano State, Nigeria. It features a broad spectrum of real-world conditions, including:

a) Fully visible license plates with minimal visual noise and standard formatting.

b) Plates affected by occlusion, low resolution, motion blur, excessive glare, or poor lighting.

c) Images containing multiple license plates, typically from dense traffic or parking scenarios.

d) Non-standard or irregular plate formats as often seen on informal or unregulated vehicles.

These images were captured using consumer-grade smartphone cameras at various times of day and weather conditions, samples are shown in Fig. 3. Importantly, no synthetic augmentation or artificial preprocessing was applied, preserving the authentic challenges inherent to LPR in emerging transportation environments.

### *B. Model Selection and Inference Setup*

To evaluate the zero-shot learning performance of modern VLMs, we adopted the Visual Prompting Playground, a web-based inference platform developed by Roboflow. This tool allows for the testing of multiple VLMs in real-time on custom visual inputs via natural language prompts. Due to platform constraints, each image had to be reduced to below 1.8 MB. Online image compression tools, namely Online Image Compressor and Image Resizer, were used to minimize image without significantly compromising visual quality.

Five VLMs were selected for evaluation primarily due to their recognition and the strong performance claims made by their respective developers:

a) GPT-4o (OpenAI)

b) Claude 4 Sonnet (Anthropic)

c) Gemini 2.0 Flash Exp (Google DeepMind)

d) Llama 3.2 Vision 90b (Meta)

e) Qwen2.5-VL-7B-Instruct (Alibaba)

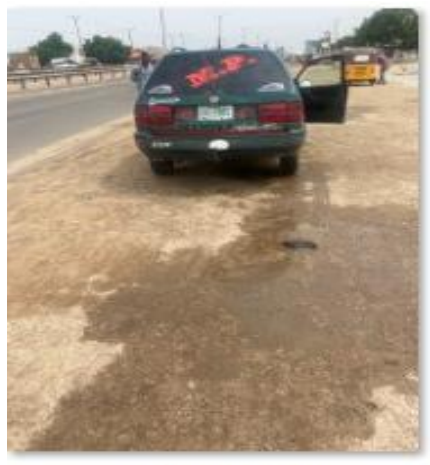

PLATE_1.JPG

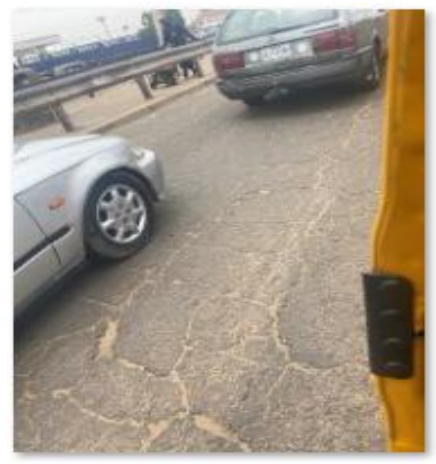

PLATE_2.JPG

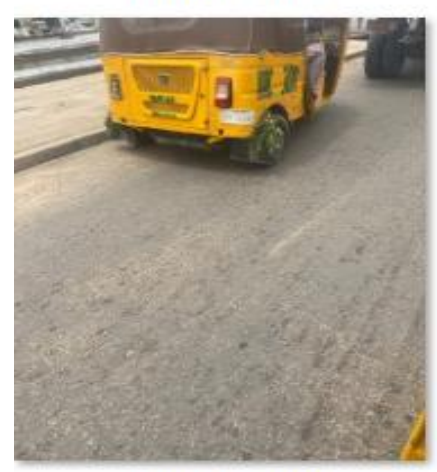

PLATE_3.JPG

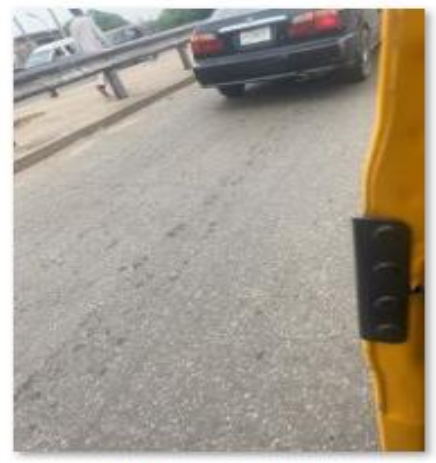

PLATE_4.JPG

Fig. 3. Samples of Captured License Plate

Each model was evaluated on the same set of 88 sample images using a **single unified prompt**, crafted to reflect the dual-task objective of plate detection and text extraction in the Nigerian context. All models were tested in parallel, in a controlled session, ensuring uniform conditions across runs.

Prompt:

> *“””Given an image containing one or more Nigerian vehicle license plates, locate each plate and extract its alphanumeric text. Each license plate number follows the format of three letters, three numbers, and two letters (e.g., ABC123XY), but some plates may be partially visible due to occlusions, damage, or poor quality. For each plate, extract as many characters as possible, even if incomplete, and return all extracted texts as a comma-separated string, with no additional text. If a plate’s text is partial, include the visible characters in the output. If no plates are found, return an empty string. Example: For an image with two plates, one fully visible as "LAG123AB" and one partially visible as "KAN45", output: LAG123AB,KAN45”””*

This prompt was carefully crafted to ensure clarity and consistency across models.

### *C. Prediction Normalization*

Due to variations in how different VLMs output their predictions, a standard normalization protocol was applied to extract consistent results:

a) Hyphens ( - ) and spaces were removed from outputs.

b) Only alphanumeric content resembling plate formats was retained.

c) Textual descriptions, explanations, or irrelevant outputs were discarded.

d) If no license plate content was found, a placeholder (- ) was recorded.

e) Inference delays were tolerated fully, ensuring each model had adequate time to process each image without penalization.

This normalization ensured fair and unified evaluation across all models, mitigating biases introduced by formatting differences.

### *D. Perfomance Evaluation Using Character Error Rate(CER)*

To quantify model performance, we adopted CER, a widely used metric in OCR and text recognition tasks. CER calculates the normalized number of character-level insertions, deletions, and substitutions needed to transform a predicted string into the ground truth.

The formula for CER is:

CER = S+D+I/N

Where:

a) S = number of substitutions

b) D = number of deletions

c) I = number of insertions

d) N = number of characters in the ground truth text

A CER of 0 indicates a perfect match, while a CER $\geq 1$ signifies a complete error. The higher the CER, the less accurate the prediction.

A Python script was developed to automatically compute CER values for each image-model pair using this formula, allowing for precise and scalable performance comparisons.

## IV. RESULTS

To evaluate the performance of the VLMs, we computed the CER for each prediction. CER captures the number of insertions, deletions, and substitutions required to transform a predicted license plate string into the ground truth, normalized by the length of the ground truth.

Table I summarizes the average CER for each VLM across the samples:

TABLE I. AVERAGE CER OF THE VLMS

| ***VLMs*** | ***Average CER*** |
|---|---|
| Gemini 2.0 Flash Exp (Google DeepMind) | 0.243 |
| Qwen2.5-VL-7B-Instruct (Alibaba) | 0.322 |
| GPT-4o (OpenAI) | 0.556 |
| Claude 4 Sonnet (Anthropic) | 0.682 |
| Llama 3.2 Vision 90b (Meta) | 0.756 |

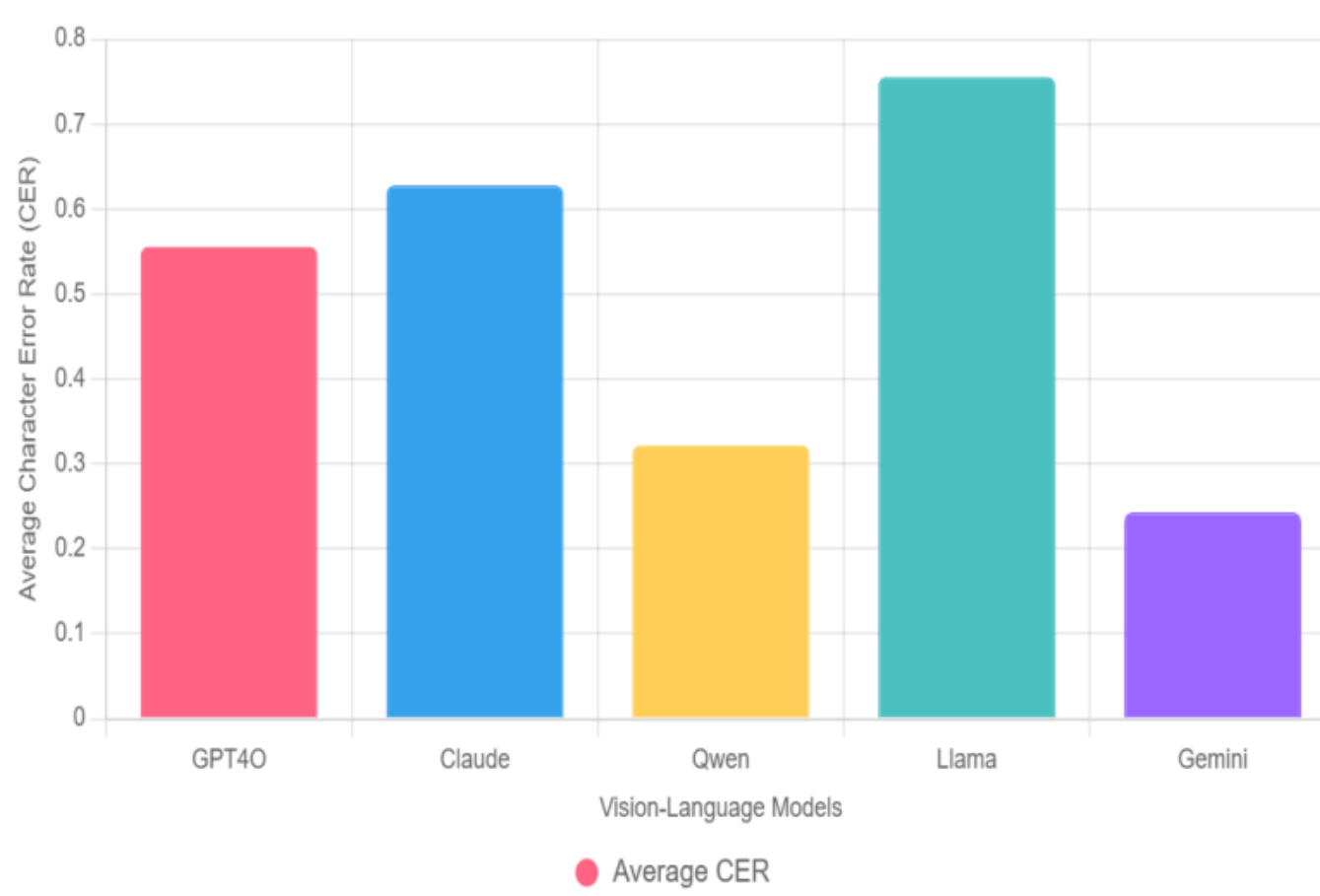

Fig. 4. Bar Chart of the Average CER

Among the evaluated models, Gemini achieved the lowest average CER (0.243), demonstrating the most accurate performance. Qwen followed closely with a CER of 0.322, indicating reliable detection and OCR capabilities. In contrast, LLaMA recorded the highest CER (0.756), suggesting notable limitations in visual-text recognition for this domain. Fig 4. Shows a bar chart of the result analysis.

## A. *Perfect Recognition (CER=0)*

TABLE II presents the number of times where each VLM achieved perfect recognition of the sample number that is CER equals to zero.

TABLE II. PERFECT RECOGNITION

| *VLMs* | *Samples* |
|---|---|
| Gemini 2.0 Flash Exp (Google DeepMind) | 37 |
| Qwen2.5-VL-7B-Instruct (Alibaba) | 30 |
| GPT-4o (OpenAI) | 30 |
| Claude 4 Sonnet (Anthropic) | 21 |
| Llama 3.2 Vision 90b (Meta) | 13 |

Only two samples (86 and 84) shown in Fig 5. were correctly recognized by all five VLMs. These plates featured clear formats, no obstructions, and minimal background noise, contributing to their perfect recognizability.

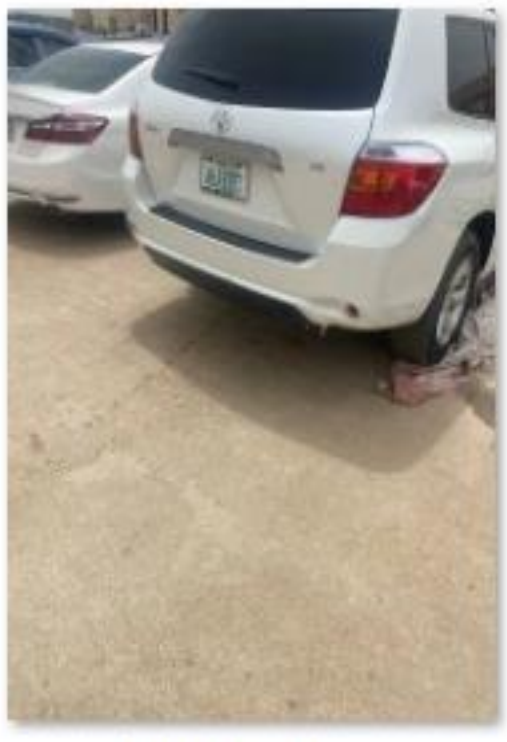

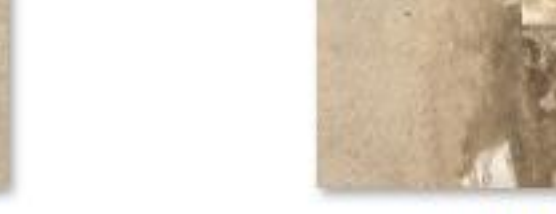

Fig. 5. Samples Correctly Recognised by all VLMs

## B. *Complete Failures (CER>=1)*

This subsection identifies how many times each model completely failed to recognize a license plate (CER $\geq$ 1) as shown in TABLE III.

Three plate samples (Plate Nos. 44, 53, and 69) were not recognized by any of the models, resulting in CER $\geq$ 1 across the board. These specific failures were due to various challenging conditions:

a) No. 44: Two plates, one blurred, one very small in the image.

b) No. 53: A single, dirty, and partially worn plate.

c) No. 69: Dark, angled plate with poor lighting.

TABLE III. COMPLETE FAILURE

| *VLMs* | *Samples* |
|---|---|
| Llama 3.2 Vision 90b (Meta) | 55 |
| Claude 4 Sonnet (Anthropic) | 37 |
| GPT-4o (OpenAI) | 29 |
| Gemini 2.0 Flash Exp (Google DeepMind) | 6 |
| Qwen2.5-VL-7B-Instruct (Alibaba) | 5 |

## C. *Handling Multiple License Plates*

A total of 28 images contained more than one license plate, making them inherently more challenging. For a model to achieve a CER of 0 in these cases, all plates in the image must be correctly predicted, which is very rare to happen. TABLE IV shows the results on samples with multiple plates. Plate No. 75 contained the highest number of plates in a single image, making it an extreme test case. Gemini 2.0 Flash Exp (Google DeepMind) again demonstrated superior capacity by achieving the lowest CER (0.014) on this image, reinforcing its robustness in multi-license plate scenarios.

TABLE IV. RESULT OF SAMPLES WITH MULTIPLE PLATES

| *VLMs* | *Average CER* | *Perfect Prediction* | *Samples Predicted Correctly* |
|---|---|---|---|
| Gemini 2.0 Flash Exp (Google DeepMind) | 0.308 | 3 | 26,37,85 |
| Qwen2.5-VL-7B-Instruct (Alibaba) | 0.477 | 2 | 58,87 |
| GPT-4o (OpenAI) | 0.756 | 1 | 58 |
| Claude 4 Sonnet (Anthropic) | 0.839 | 0 | - |
| Llama 3.2 Vision 90b (Meta) | 0.869 | 0 | - |

## V. DISCUSSION

### A. *Performance Insights*

This study reveals that Gemini 2.0 Flash Exp (Google DeepMind) and Qwen2.5-VL-7B-Instruct (Alibaba) significantly outperform other VLMs in accurately identifying Nigerian license plate numbers, a task that combines both detection and text extraction [17][18]. These two models achieved lower average CER, fewer total errors, and higher perfect recognition counts, indicating superior performance.

Their consistent accuracy suggests that Gemini 2.0 Flash Exp (Google DeepMind) and Qwen2.5-VL-7B-Instruct (Alibaba) likely benefit from better multi-modal pre-training strategies or more fine-grained visual-textual alignment mechanisms [19]. Gemini 2.0 Flash Exp (Google DeepMind), in particular, demonstrated strong performance even in challenging cases involving multiple plates, occlusions, and varying plate conditions, making it highly suitable for real-world deployment in traffic surveillance and security applications [20].

On the other hand, LLaMA 3.2 Vision 90B (Meta) recorded the highest error rate and the lowest number of perfect recognitions, highlighting limitations in its visual-text understanding capabilities. Despite the general strength in other NLP tasks, its performance in this setting was moderate. A closer look at its predictions reveals that LLaMA was able to detect license plates but often failed to extract text correctly, frequently outputting responses such as *"I can't extract the text from the plate"*. This indicates a clear gap in its extraction ability, despite being able to localize the plate region [21].

### B. *VLMs over YOLO+OCR Approach*

As stated earlier, Traditional license plate recognition systems typically rely on a multi stage pipeline: first, an object detection model such as YOLO is used to locate and crop license plates from an image, and then an OCR model is applied to extract the text. While this approach has proven effective over time, it introduces significant limitations. Each stage of the YOLO+OCR pipeline introduces its own complexities and potential points of failure [4]. Errors in detection, such as incorrect or partial bounding boxes, often lead to poor OCR results downstream. Additionally, integrating and maintaining these stages requires more computational resources, increases latency, and adds engineering overhead [22].

Another major limitation of the YOLO+OCR method is its heavy reliance on large annotated datasets. Developing high-performing models in this pipeline demands thousands of labeled images with bounding boxes and corresponding plate texts. In many local contexts, such datasets are scarce or completely unavailable, making training or fine-tuning difficult without costly data collection. It also requires frequent monitoring and retraining to stay effective in dynamic real-world environments, where plate styles, lighting, image quality, and viewing angles may frequently change. This need for continual supervision and manual intervention makes the approach less scalable, especially in low-resource settings [16].

On the other hand, VLMs offer a more elegant and unified solution. This study evaluated pre-trained VLMs on challenging Nigerian license plate images in a zero-shot learning setting, that is, without any additional training or fine-tuning. Despite this, models like Gemini 2.0 Flash Exp (Google DeepMind) and Qwen2.5-VL-7B-Instruct (Alibaba) delivered impressive results. These models were able to identify and extract license plate text directly from raw images, combining visual recognition and text generation in a single step [23]. This eliminates the need for bounding box annotations, separate detection modules, or OCR engines.

The strength of VLMs lies in their ability to generalize well across unseen domains. Without ever being trained on Nigerian license plates, they demonstrated strong performance. Their adaptability and low CER highlight their robust multimodal alignment, likely due to diverse and large-scale pre-training [24].

### C. *Comparative Analysis of the Five VLMs with Developers' Claims*

An in-depth evaluation of five evaluated VLMs, reveals significant variation in performance, particularly under real-world conditions involving blur, occlusion, and multi-plate inputs [17]. Each model was assessed using three key metrics: average CER, number of perfect recognitions (CER = 0), and number of complete failures (CER ≥ 1). Below is a breakdown of each model's performance alongside an assessment of whether it meets its developers' claims.

**Gemini 2.0 Flash Exp (Google DeepMind)** emerged as the top performer, achieving the lowest average CER (0.243), the highest number of perfect recognitions (37), and one of the fewest complete failures (6). It demonstrated exceptional robustness in complex scenarios. These results strongly validate Google's claims of Gemini being a state-of-the-art multimodal model with advanced vision and reasoning capabilities. The model delivered high accuracy and required minimal tuning, confirming its advertised reliability in real-world tasks.

**Qwen2.5-VL-7B-Instruct (Alibaba)** followed closely, posting an impressive average CER of (0.322), (30) perfect recognitions, and only (5) complete failures, the lowest among all models. It showed consistent performance and handled visual complexity with high precision. This outcome supports claims from Chinese research communities that emphasize Qwen's fine-grained vision-text alignment and strong cross-modal understanding. This VLM is not widely promoted globally; its performance in this study positions it as an underrated but highly capable model in practical applications.

**GPT-4o (OpenAI)** delivered moderate performance with an average CER of (0.556) and (21) perfect recognitions, indicating reasonable but not exceptional OCR accuracy. The model struggled particularly with text extraction, suggesting that its strength lies more in general reasoning than in specialized detection and OCR tasks. OpenAI markets GPT-4o as a highly capable vision-language model, this study reveals a gap between those claims and its practical effectiveness in detection and OCR contexts, highlighting the need for task-specific fine-tuning to unlock its full potential.

**Claude 4 Sonnet (Anthropic)**, recorded an average CER of 0.682, matched GPT-4o in perfect recognitions (30), but had a much higher rate of complete failures (37). Although positioned by Anthropic as a high-performing multimodal model with an emphasis on safety and broad generalization, its performance in this evaluation indicates limitations in specific context especially with handling noisy, localized visual data. The results suggest that Claude 4 Sonnet may lack the visual diversity or OCR-specific pre-training needed for robust text recognition in real-world scenarios, falling short of its marketed strengths in this domain.

**Llama 3.2 Vision 90b (Meta)**, ranked lowest in performance, with the highest average CER (0.756), the fewest perfect recognitions (13), and the most complete failures (55). In many cases, the model correctly identified the presence of a license plate but failed to extract any meaningful text, reflecting a weakness in vision-text fusion. This underperformance aligns with broader concerns that LLaMA, being more language-centric and less extensively trained on visual data, lacks the multimodal robustness required for OCR. Consequently, its results in this study do not support Meta's claims of advanced visual understanding.

While most model developers claim state-of-the-art status for their VLMs based on popular benchmarks such as VQAv2 or TextVQA, this evaluation highlights a disconnect between those benchmarks and actual performance in real-world, localized tasks like Nigerian license plate recognition. **Gemini 2.0 Flash Exp** and **Qwen2.5-VL-7B-Instruct** stand out for aligning with and even exceeding their advertised capabilities, whereas **GPT-4o**, **Claude 4 Sonnet**, and **LLaMA 3.2 Vision** underperformed, revealing the importance of OCR-specific pretraining and diverse visual datasets for practical deployment [25][26].

## VI. Conclusion

This study presents the comparative evaluation of selected VLMs, Gemini 2.0 Flash Exp (Google DeepMind), Qwen2.5-VL-7B-Instruct (Alibaba), GPT-4o (OpenAI), Claude 4 Sonnet (Anthropic), and Llama 3.2 Vision 90b (Meta), on the task of recognizing Nigerian license plates. Through the use of CER as the primary metric, we assessed model performance across both single and multiple plate images, offering a holistic perspective on accuracy, reliability, and robustness.

Our findings show that Gemini 2.0 Flash Exp (Google DeepMind) and Qwen2.5-VL-7B-Instruct (Alibaba) significantly outperform other models in terms of accuracy, resilience to challenging conditions, and handling multi-plate inputs, despite being evaluated in a zero-shot setting with no additional fine-tuning. This demonstrates the strong visual-textual alignment and practical readiness of these models for localized tasks. Gemini 2.0 Flash Exp (Google DeepMind), in particular, achieved the lowest average CER and the highest number of perfect recognitions.

Conversely, models like Claude 4 Sonnet (Anthropic) and Llama 3.2 Vision 90b (Meta) underperformed, with high error rates and numerous complete failures, suggesting that despite their general capabilities or popularity in benchmarks, they may not be well-suited for specialized OCR tasks without targeted adaptation. GPT-4o (OpenAI), though stronger than Claude and LLaMA, showed some limitations in detection and text extraction despite its strong general reasoning ability.

Furthermore, this study challenges the reliability of leaderboard-based SOTA benchmarks as universal indicators of real-world performance. It also highlights the efficiency and potential of VLMs over traditional YOLO+OCR pipelines by eliminating the need for manual data collection, retraining, or multi-step architectures.